\documentclass{article}

\usepackage{PRIMEarxiv}

\usepackage[utf8]{inputenc} 
\usepackage[T1]{fontenc}    
\usepackage{hyperref}       
\usepackage{url}            
\usepackage{booktabs}       
\usepackage{multirow}
\usepackage{amsfonts}       
\usepackage{amsmath}
\usepackage{nicefrac}       
\usepackage{subfig}
\usepackage{microtype}      
\usepackage{lipsum}
\usepackage{fancyhdr}       
\usepackage{graphicx}       
\graphicspath{{media/}}     

\title{Conformal Segmentation in Industrial Surface Defect Detection with Statistical Guarantees}

\author{
  Cheng Shen, Yuewei Liu \\
  School of Information Science, School of Mathematics and Statistics\\
  Lanzhou University\\
  Lanzhou\\
  \texttt{320220950671@lzu.edu.cn, lyw@lzu.edu.cn} \\
}

\begin{document}
\maketitle

\begin{abstract}
In industrial settings, surface defects on steel can significantly compromise its service life and elevate potential safety risks. 
Traditional defect detection methods predominantly rely on manual inspection,
which suffers from low efficiency and high costs. 
Although automated defect detection approaches based on Convolutional Neural Networks(e.g., Mask R-CNN) have advanced rapidly, 
their reliability remains challenged due to data annotation uncertainties during deep model training and overfitting issues. These limitations may lead to detection deviations when processing the given new test samples, rendering automated detection processes unreliable.
To address this challenge, we first evaluate the detection model's practical performance through calibration data that satisfies the independent and identically distributed (i.i.d) condition with test data. Specifically, we define a loss function for each calibration sample to quantify detection error rates, such as the complement of recall rate and false discovery rate. Subsequently, we derive a statistically rigorous threshold based on a user-defined risk level to identify high-probability defective pixels in test images, thereby constructing prediction sets (e.g., defect regions). This methodology ensures that the expected error rate (mean error rate) on the test set remains strictly bounced by the predefined risk level.
Additionally, we observe a negative correlation between the average prediction set size and the risk level on the test set, establishing a statistically rigorous metric for assessing detection model uncertainty. Furthermore, our study demonstrates robust and efficient control over the expected test set error rate across varying calibration-to-test partitioning ratios, validating the method's adaptability and operational effectiveness.
\end{abstract}

\keywords{Surface Defect Detection; Mask R-CNN;Uncertainty Quantification; Statistical Guarantees; False Discovery Rate; False Negative Rate; Risk Control}

\section{Introduction}
In modern industrial production, metallic structural materials are susceptible to surface defects such as cracks, pores, and folds during manufacturing, transportation, and service processed due to multiple factors including mechanical stress and environmental corrosion. According to an ASTM (American Society for Testing and Materials) investigation report, approximately 62\% of mechanical component failures can be attributed to stress concentration effects induced by surface defects. Such defects not only significantly reduce material fatigue life but may also trigger cascading structural failures - for instance, 0.1mm-level cracks in pressure vessels can propagate under cyclic loading, potentially leading to catastrophic bursting accidents. Therefore, comprehensive and precise defect detection is imperative.

Traditional surface defect inspection primarily relies on manual visual examination, which suffers from strong subjectivity and low detection efficiency. Studies indicate that human inspection achieves an average FDR (False Detection Rate) of 18.7\% in complex industrial scenarios\cite{liu2021defect}. Recent advancements in deep learning algorithms, particularly Mask R-CNN, have demonstrated remarkable progress in defect segmentation tasks, achieving over 94\% detection accuracy through end-to-end feature learning\cite{wang2020deep}. However, these black-box models exhibit overconfident predictions and may generate severe misjudgments when encountering unseen defect types or noise interference.

To address these trustworthiness challenges~\cite{xiao2023towards,gawlikowski2023survey,wang2025word}, this study employs the Conformal Prediction (CP) framework, which constructs asymmetric prediction confidence intervals through calibration sets to provide statistically verifiable coverage guarantees under arbitrary data distributions. While conventional CP methods ensure theoretical lower bounds for defect detection rates, they fail to effectively control false detection proportions. We therefore propose a risk-controlled enhancement framework by defining FDR (False Discovery Rate) as the controllable risk metric:

\begin{equation}
    \text{FDR} = \mathbb{E}\left[\frac{\text{False Defect Count}}{\text{Total Detected D}}\right].
\end{equation}

This framework adaptively optimizes model sensitivity while maintaining precision constraints, ultimately achieving a balance between reliability and detection efficiency.

Preliminary experiments with different backbone networks evaluated detection performance, with results shown in Table \ref{tab: preexperiment_results} revealing the inherent unreliability of deep learning-based defect detection methods. Our subsequent validation demonstrates consistent FDR control performance across datasets, indicating superior detection accuracy and reduced false positives. By designing a loss function reflecting false negative proportions for each calibration sample, we achieve user-specified FNR (False Negative Rate) metrics on test sets. Comparative analysis of FNR control performance under varying data splitting ratios systematically examines calibration set sensitivity, confirming strict adherence to reliability constraints across all partitioning schemes, thereby verifying method robustness.

\section{Related Works}

This section reviews prior work in surface defect detection, categorized into traditional machine vision approaches, deep learning-based methods and methods combined with CP and CRC.

\subsection{Traditional Machine Vision-Based Defect Detection Methods}
Traditional machine vision techniques for defect detection often rely on analyzing local anomalies, matching templates, or classifying handcrafted features. These methods typically exploit statistical or structural properties of images.

One prominent approach involves statistical texture analysis to identify local deviations. For instance, Liu et al. \cite{liu2017steel}
proposed an unsupervised method employing a Haar-Weibull variance model to represent the texture distribution within local image patches.
This statistical modeling allowed for the detection of defects by identifying areas whose texture significantly deviated from the learned normal patterns, offering an effective way to characterize surface anomalies without prior defect knowledge.

Another common strategy is template matching, where test images are compared against defect-free reference templates. To address challenges like rotation and scale variations, Chu and Gong \cite{chu2015invariant}
developed an invariant feature extraction method based on smoothed local binary patterns (SLBP) combined with statistical texture features.
Their contribution lies in achieving robustness to geometric transformations common in industrial settings, enabling accurate defect classification even when the defect's orientation or size varies relative to the template.

\subsection{Deep Learning-Based Defect Detection Methods}
Deep learning, particularly Convolutional Neural Networks (CNNs), has demonstrated superior performance in handling the complexity and variability of defects found in industrial environments, largely due to their ability to automatically learn hierarchical features directly from data. Methods enabling pixel-level localization, such as segmentation or salient object detection, are crucial for precise defect analysis.

One line of research focuses on developing specialized network architectures for accurate pixel-level defect identification. Song et al. \cite{song2020edrnet}
proposed the Encoder-Decoder Residual Network (EDRNet) specifically for salient object detection of surface defects on strip steel. Their architecture incorporates residual refinement structures to enhance feature representation and boundary localization.
This work provides an application-specific deep learning solution for generating pixel-accurate saliency maps highlighting defect regions, which is essential for detailed defect assessment and downstream analysis in steel manufacturing.

For applications requiring precise defect boundaries and potentially instance differentiation, segmentation networks are employed. Huang et al. \cite{huang2021automatic}
proposed a Deep Separable U-Net, which utilizes depth-wise separable convolutions within a U-shape encoder-decoder architecture combined with multi-scale feature fusion.
This work focused on optimizing the trade-off between segmentation accuracy and computational efficiency. By leveraging lightweight convolutions and effective feature fusion, they achieved competitive performance suitable for automatic, pixel-level defect segmentation in resource-aware industrial scenarios.

\subsection{Conformal Prediction and Conformal Risk Control}
Existing surface defect detection methods, whether traditional machine vision or standard deep learning, often lack the strict reliability guarantees vital for industrial settings; traditional techniques struggle with complexity, while deep learning can be overconfident and uncontrolled regarding critical error rates. Conformal Prediction (CP) \cite{vovk1999machine} offers a robust alternative via distribution-free, model-agnostic prediction sets with guaranteed coverage ((\(1 - \alpha\)). However, coverage guarantees alone are often inadequate, and sets can be too conservative. Conformal Risk Control (CRC) \cite{angelopoulos2022conformal} advances CP's statistically sound framework by enabling direct control over the expected value of customizable, task-specific loss functions (e.g., FDR, FNR), ensuring they remain below a pre-set risk level (\(\alpha\)).

The following subsections delve into specific prior works that leverage these conformal methodologies.

\subsubsection{Conformal Prediction}
Conformal prediction can provide statistically rigorous guarantees of task-specific metrics by constructing prediction sets~\cite{angelopoulos2022conformal,wang2024conu,hulsman2024conformal,wang2025sample,wang2025sconu}. 
To address the need for reliable diagnostic predictions, Zhan et al. (2020) \cite{zhan2020electronic} implemented k-NN based Conformal Prediction (CP) for an electronic nose system detecting lung cancer. By leveraging a nonconformity score derived from nearest neighbor distances, they generated prediction sets and associated uncertainty metrics (confidence, credibility), empirically validating the key theoretical guarantee of CP, controlling the error rate below a predefined significance level \(\alpha\), particularly within an online prediction protocol.

Integrating uncertainty quantification into real-time visual anomaly detection, Saboury and Uyguroglu (2025) \cite{saboury2025uncertainty} utilize the Conformal Prediction (CP) framework atop an unsupervised autoencoder. They calculate a nonconformity score for each image based on its reconstruction error (MAE + MSE). Using a calibration set of normal images, they compute statistically valid p-values for test images according to the standard inductive CP procedure. This allows classifying an image as anomalous if its p-value falls below a chosen significance level \(\alpha\), with the key guarantee that the false positive rate on normal images is controlled at or below \(\alpha\).

\subsubsection{Conformal Risk Control}
Conformal Risk Control (CRC) represents a generalization of the principles underlying Conformal Prediction, moving beyond the typical goal of coverage guarantees \cite{angelopoulos2022conformal}. Instead of solely controlling the probability of miscoverage (a specific 0/1 risk), CRC provides a framework for controlling the expected value of more complex, user-specified loss functions. This allows for tailoring the uncertainty quantification to specific application needs where different types of errors might have varying costs or consequences, offering guarantees that the chosen risk metric will be bounded, often under the same minimal assumption of data exchangeability used in CP. While standard Conformal Prediction often concentrates on correctness coverage, CRC offers a pathway to manage broader notions of predictive risk using conformal methods.

Extending conformal methods to semantic segmentation, Mossina et al. (2023) \cite{mossina2024conformal} utilized Conformal Risk Control (CRC) to produce statistically valid multi-labeled segmentation masks. They parameterized prediction sets using the Least Ambiguous Set-Valued Classifier (LAC) approach based on a threshold \(\lambda\) applied to pixel-wise softmax scores. By defining specific monotonic loss functions, such as binary coverage or pixel miscoverage rate, and employing the CRC calibration procedure, they determined the optimal threshold \(\hat{\lambda}\) ensuring the expected value of the chosen loss on unseen images is bounded by the target risk \(\alpha\).

To enhance the reliability of object detectors in safety-critical railway applications, Andéol et al.~\cite{andeol2023confident} implemented an image-wise Conformal Risk Control (CRC) approach. They defined specific loss functions suitable for object detection, including a box-wise recall loss (proportion of missed objects) and a pixel-wise recall loss (average proportion of missed object area). By computing these losses on a calibration set and applying the CRC methodology, they determined the necessary bounding box adjustments (\(\hat{\lambda}\)) to ensure that the expected value of the chosen image-level risk (e.g., missed pixel area) is controlled below a predefined threshold \(\alpha\) on new data.

To address confidence calibration issues in medical segmentation, Dai et al.~\cite{dai2025statistical} leveraged Conformal Risk Control (CRC). Their key contribution was the design of specific loss functions, calculated per calibration sample, that directly corresponded to 1-Precision (for FDR control) or 1-Recall (for FNR control) at a given segmentation threshold. Applying the CRC methodology allowed them to determine a data-driven segmentation threshold \(\hat{\lambda}\) that guarantees the expected value of either the chosen FDR or FNR metric on unseen test images is controlled below a pre-specified risk tolerance \(\alpha\).

Addressing the need for reliable uncertainty quantification in AI-assisted lung cancer screening, Hulsman et al. (2024) \cite{hulsman2024conformal} employed Conformal Risk Control (CRC). They defined the per-scan False Negative Rate (FNR) as the risk metric of interest and used a calibration dataset to find the appropriate confidence threshold \(\lambda\) for their nodule detection model. To be more detailed, by defining a loss function based on the per-scan False Negative Rate and applying the CRC framework to calibrate the detector's confidence threshold, they achieved guaranteed control over the expected sensitivity (as 1-FNR) for new patient scans, demonstrating CRC's utility in safety-critical medical imaging applications. This CRC procedure guarantees that the expected FNR on a future scan will not exceed a pre-specified risk tolerance \(\alpha\), offering a principled method to ensure high sensitivity with statistical validity.

\section{Risk Control Framework}
\subsection{Mask R-CNN}
The mainstream instance segmentation method, Mask R-CNN \cite{he2017mask}, integrates object detection and segmentation by extending Faster R-CNN with a pixel-level segmentation branch. The model can be formalized as the following optimization problem.

\begin{equation}
    \mathcal{M}(x) = \{ (B_i, C_i, M_i) \}_{i=1}^K = \arg\max_{\theta} \sum_{(x,y)\in\mathcal{D}} \mathcal{L}_{\text{multi}}(f_\theta(x), y).
\end{equation}
where the multi-task loss function is defined as:

\begin{equation}
    \mathcal{L}_{\text{multi}} = \lambda_1\mathcal{L}_{\text{cls}} + \lambda_2\mathcal{L}_{\text{box}} + \lambda_3\mathcal{L}_{\text{mask}}.
\label{eq: multi}
\end{equation}

The hyperparameters \(\lambda_i\) balance classification loss, bounding box regression loss, and mask segmentation loss. Despite its strong performance on benchmark datasets, Mask R-CNN lacks uncertainty quantification, which may lead to dangerously overconfident false predictions when test data deviates from the training distribution.

For an input image \(x_i\), the model outputs three semantic components:
\begin{itemize}
    \item Bounding box coordinates \(B_i = (x_{min}, \, y_{min}, \, x_{max}, \, y_{max})\)
    \item Defect category label \(C_i \in \{1,...,K\}\), where K denotes the predefined number of classes
    \item Binary segmentation mask \(M_i \in \{0,1\}^{h \times w}\) that precisely localizes defect pixels
\end{itemize}

\subsection{Rules of Conformal Prediction and Conformal Risk Control}
Conformal Prediction (CP) provides a framework to address the reliability challenges in machine learning models by offering rigorous statistical guarantees, particularly valuable for deep learning systems \cite{vovk1999machine}. Its core contribution is constructing prediction sets that provably contain the true outcome with at least a user-specified probability (\(1 - \alpha\)), known as marginal coverage. This guarantee holds under minimal assumptions, as CP is distribution-free and model-agnostic, making it widely applicable across diverse data types and complex black-box models. Operationally, it leverages an independent calibration dataset to compute nonconformity scores, determines a critical threshold based on the quantiles of these scores, and uses this threshold to form prediction sets for new instances, all without needing specific distributional knowledge. The complete framework of Conformal Prediction is illustrated in previous work \cite{vovk1999machine}.

While powerful, traditional Conformal Prediction is limited primarily to guaranteeing coverage probability, often failing to provide assurances for specific task-relevant metrics (like false discovery rates) and sometimes producing prediction sets too large for practical utility. Conformal Risk Control (CRC) was developed to address these shortcomings by extending CP's principles. CRC inherits the statistical rigor of CP but shifts the focus from mere coverage to controlling the expected value of a user-defined, measurable loss function (\(ell(\cdot\)) pertinent to the application (e.g., controlling the expected False Discovery Rate). By ensuring this expected risk satisfies (\(\mathbb{E}[\ell] \leq \alpha\)) for a chosen risk level (\(\alpha\)), CRC provides a more flexible and generalized framework capable of accommodating diverse application requirements where specific types of errors need explicit management.

The Conformal Risk Control theory is described as follows \cite{angelopoulos2022conformal}:

Given an i.i.d calibration dataset \( \{(x_i, y^*_i)\}^n_{i=1} \) where \(x_i \in \mathbb{R}^{h\times w}\) denotes input images and \(y^*_i \subseteq \{1,...,h\} \times \{1,...,w\}\) represents defect pixel coordinates, consider a deep learning model \(f: \mathbb{R}^{h \times w} \to [0, 1]^{h \times w}\) that outputs pixel-wise probability maps. Define a threshold-controlled prediction set:

\begin{equation}
    C_i(\lambda) = \{(j,k) | f(x_i)_{j,k} \geq 1-\lambda\}.
\label{eq: predictionset}
\end{equation}

This set contains all pixels with prediction confidence exceeding \(1-\lambda\). For each calibration sample, we construct a FDR (False Discovery Rate) loss function:

\begin{equation}
    l_i(\lambda) = 1 - \frac{|C_i(\lambda)\cap y_i^*|}{|C_i(\lambda)| \vee 1} \leq 1.
\label{eq: l_i}
\end{equation}

where \(|\cdot|\) denotes set cardinality and \(\vee\) prevents division-by-zero errors for empty prediction sets. The loss function monotonically decreases with \(\lambda\), reflecting reduced false discovery rates under higher confidence thresholds.

The empirical risk is computed on the calibration set:
\begin{equation}
    L_n(\lambda) = \frac{1}{n}\sum_{i=1}^n l_i(\lambda).
\label{eq: empirical risk}
\end{equation}

Under the exchangeability assumption, the expected test risk satisfies:

\begin{equation}
    \mathbb{E}[l_{n+1}(\lambda)] = \frac{nL_n(\lambda) + [l_{n+1}(\lambda)]}{n+1}.
\label{eq: conformal risk control}
\end{equation}

The optimal threshold is determined by solving:

\begin{equation}
\widehat{\lambda}=inf\left\{\lambda{:}\frac{nL_{n}(\lambda)+1}{n+1}\leq\alpha\right\}=inf\left\{\lambda{:}L_{n}(\lambda)\leq\frac{\alpha(n+1)-1}{n}\right\}.
\label{eq: lambda}
\end{equation}

The threshold selection strategy ensures rigorous risk control:

\begin{equation}
\mathbb{E}\left[l_{n+1}(\widehat{\lambda})\right]=\frac{nL_{n}(\widehat{\lambda})+l_{n+1}(\widehat{\lambda})}{n+1}\leq\frac{nL_{n}(\lambda)+1}{n+1}\leq\alpha.
\label{eq: thresholdchoice}
\end{equation}

The method achieves statistically strict FDR control while dynamically adjusting thresholds to maximize prediction coverage.

\subsection{Workflow of our Approach}
To achieve reliable industrial surface defect detection with statistical guarantees, we propose a risk-controlled segmentation framework, as illustrated in Figure 1. This approach combines the instance segmentation capabilities of Mask R-CNN with the robust guarantees offered by Conformal Risk Control (CRC). We selected Mask R-CNN because of its strong baseline performance in localizing and classifying defects at the pixel level. However, its standard output lacks uncertainty quantification. To address this, CRC is employed to calibrate Mask R-CNN's predictions, specifically controlling the False Discovery Rate (FDR) — the proportion of falsely identified defect pixels — to be below a user-defined risk level, $\alpha$, thereby enhancing the reliability of the automated detection process.

The workflow begins by processing an input test image (\(X_{test}\))  using a pre-trained Mask R-CNN model. In the initial defect detection step, this generates pixel-wise defect probability maps (\(P^{test}\)), which retain crucial raw uncertainty information required for the conformalization process. Although Mask R-CNN also produces binary masks, we use the probability maps for calibration purposes. 

At the heart of our method is the CRC Calibration stage, shown in the central block of Figure 1. Using a separate calibration dataset (\(D_{calib} = \{(x_i, y_i)\}_{i=1}^n\)), where \(y_i\) denotes the ground-truth defect pixels for the image \(x_i\), we generate prediction sets \(C_\lambda(x_i)\)  for each calibration sample. These sets are based on a candidate threshold \(\lambda\) applied to the model’s probability outputs: ( \(C_\lambda(x_i) = \{(j, k) | f(x_i)_{j,k} \ge 1-\lambda\}\)). We then compute a loss function for each sample — specifically, the FDR loss \(l_i(\lambda)\), as detailed in Eq. \ref{eq: l_i}), which quantifies the false discovery proportion at that threshold. The empirical risk \(L_n(\lambda)\) (Eq. \ref{eq: empirical risk}) is calculated by averaging these losses over the calibration set. Using this empirical risk and the target risk level \(\alpha\), we determine the optimal data-driven threshold \(\hat{\lambda}\) via constrained optimization as detailed in Eq. \ref{eq: multi}), identifying the least conservative (largest) threshold that satisfies the risk constraint based on the calibration data.

Finally, the risk-calibrated threshold \(\hat{\lambda}\) is applied to generate statistically rigorous prediction sets on the test image. The final prediction set (\(S_{test}\)) is formed by including all pixels in the test probability map \(P^{test}\) whose confidence score is greater than or equal to \(1 - \hat{\lambda}\) (\(S_{test} = \{(j,k) : P^{test}_{j,k} \ge 1-\hat{\lambda}\}\)).  This entire process provides a formal guarantee as presented by Eq. \ref{eq: thresholdchoice}): the expected FDR of the resulting prediction set \(S_{test}\) on unseen test data is rigorously controlled to be no greater than the predefined risk level \(\alpha\) (\(E[\text{FDR}(S_{test})] \le \alpha\)). 

\begin{figure}[htbp]
\centering
\includegraphics[width=0.95\textwidth]{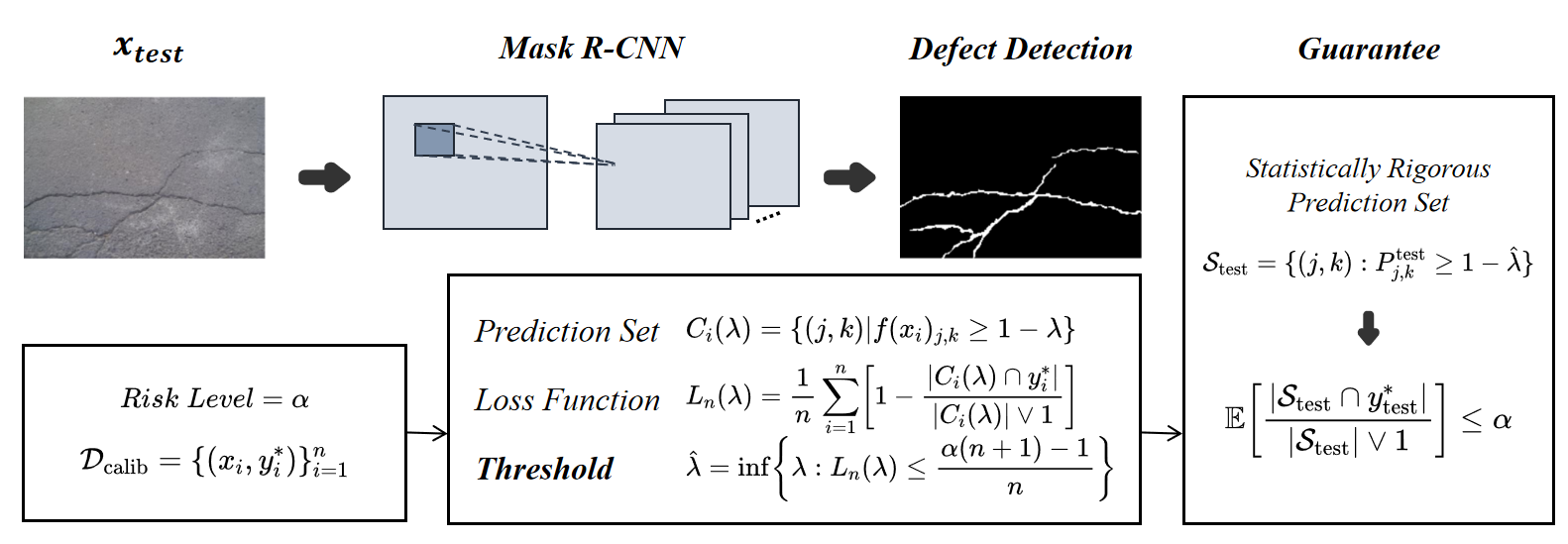}
\caption{Workflow of our Approach}
\label{fig: workflow}
\end{figure}

This framework ensures user-specified risk control while adaptively tuning detection sensitivity based on the data-dependent threshold \(\hat{\lambda}\), effectively balancing statistical rigor with operational practicality. 

\section{Experiment Analysis}
\subsection{Datasets and Benchmarks}
\begin{figure}[htbp]
    \centering
    \subfloat[Multi-scale Defect Examples from Severstal]{
        \includegraphics[width=.45\linewidth, height=.3\textheight, keepaspectratio]{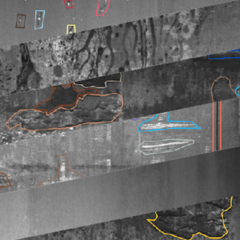}
    }
    \hfill
    \subfloat[Inter-class Similarity Analysis in NEU]{
        \includegraphics[width=.45\linewidth, height=.3\textheight, keepaspectratio]{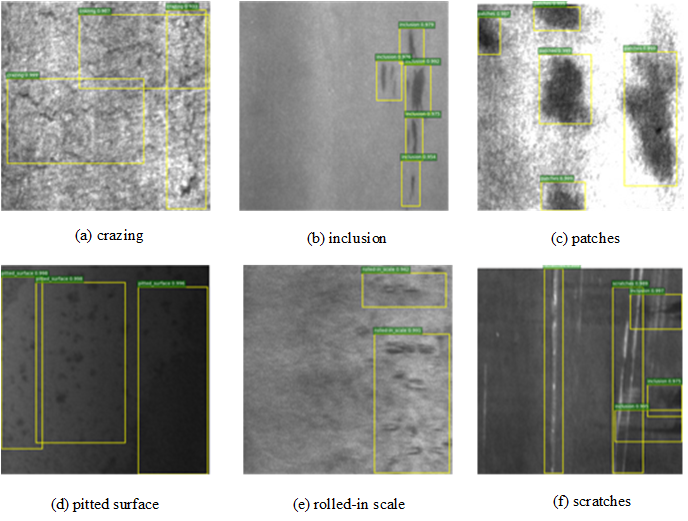}
    }
    \caption{Visual Dataset Analysis}
    \label{fig: dataset_visual}
\end{figure}

This study is validated on two complementary steel surface defect datasets, Figure~\ref{fig: dataset_visual} gives an intuitive visualization:

\begin{itemize}
    \item \textbf{Severstal Industrial Inspection Dataset}: Derived from the industrial inspection platform of Severstal, a Russian steel giant, this dataset contains 25,894 high-resolution images (2,560 \(\times\) 1,600 pixels) covering four typical defect categories in cold-rolled steel production:
    \begin{itemize}
        \item Class 1: Edge cracks (37.2\% prevalence)
        \item Class 2: Inclusions (28.5\% prevalence)
        \item Class 3: Surface scratches (19.8\% prevalence)
        \item Class 4: Rolled-in scale (14.5\% prevalence)
    \end{itemize}
    This dataset exhibits multi-scale defect characteristics under real industrial scenarios, with the smallest defect regions occupying only 0.03\% of the image area. Each image contains up to three distinct defect categories, annotated with pixel-wise segmentation masks and multi-label classifications. The data acquisition process simulates complex industrial conditions, including production line vibrations and mist interference. The primary challenge is quantified by:
    \begin{equation}
        \mathcal{C}_1 = \frac{1}{N}\sum_{i=1}^N \left( \frac{\text{Defective Pixel}_i}{\text{Total Pixels}} \right) = 0.18\%.
    \end{equation}
    where \(\mathcal{C}_1\) denotes the average defect coverage ratio, highlighting the difficulty of small-target detection.

    \item \textbf{NEU Surface Defect Benchmark}: A widely adopted academic standard comprising 1,800 grayscale images (200$\times$200 pixels) uniformly covering six hot-rolled steel defect categories:  
    \begin{itemize}
        \item Rolled-in scale (RS)
        \item Patches (Pa)
        \item Crazing (Cr) 
        \item Pitted surface (PS)
        \item Inclusion (In)
        \item Scratches (Sc)
    \end{itemize}
    The dataset provides dual annotations (bounding boxes and pixel-level masks). Its core challenge arises from the contrast between intra-class variation \(\mathcal{D}_{\text{intra}}\) and inter-class similarity \(\mathcal{D}_{\text{inter}}\): 
    \begin{equation}
        \frac{\mathcal{D}_{\text{intra}}}{\mathcal{D}_{\text{inter}}} = \frac{\mathbb{E}[d(f(x_i), f(x_j))|_{y_i=y_j}]}{\mathbb{E}[d(f(x_i), f(x_j))|_{y_i\neq y_j}]} = 1.27.
    \end{equation}
    where $f(\cdot)$ denotes a ResNet-50 feature extractor and $d(\cdot)$ measures cosine distance. A ratio exceeding 1 indicates that inter-class discriminability is weaker than intra-class variability, posing significant challenges for classifier design.
\end{itemize}

\subsection{Experimental Setup}
To evaluate the impact of different feature extraction modules, we implemented Mask R-CNN with ResNet-50 and several alternative backbone networks for feature extraction. For fair comparison, all backbones were initialized with ImageNet pretrained weights and fine-tuned under identical training protocols:

\begin{itemize}
    \item \textbf{ResNet-50}: A classic residual network comprising 50 layers (49 convolutional layers and 1 fully connected layer). It employs residul blocks with shortcut connections to mitigate gradient vanishing issues and optimizes computational effieciency through bottleneck design.\cite{he2016deep}

    \item \textbf{ResNet-34}: A lightweight variant of ResNet with 34 convolutional layers. It stacks BasicBlocks (dual 3×3 convolutions) to reduce computational complexity, making it suitable for low-resource scenarios. \cite{he2016deep}  
  
    \item \textbf{SqueezeNet}: Utilizes reduced 3×3 convolution kernels and "Fire modules" (squeeze + expand layers) to compress parameters to 1.2M. While achieving lower single-precision performance than standard ResNets, it is ideal for lightweight applications with relaxed accuracy requirements. \cite{iandola2016squeezenet}  
  
    \item \textbf{ShuffleNetv2\_x2}: Addresses feature isolation in group convolutions via channel shuffle operations. The 2× width factor version balances 0.6G FLOPs computational cost with 73.7\% accuracy through depthwise separable convolutions and memory access optimization. \cite{ma2018shufflenet}  
  
    \item \textbf{MobileNetv3-Large}: Combines inverted residual structures, depthwise separable convolutions, and squeeze-excitation (SE) modules for efficient feature extraction. Enhanced by h-swish activation, it achieves 75.2\% accuracy. \cite{howard2019searching}  
  
    \item \textbf{GhostNetv3}: Generates redundant feature maps via cheap linear operations, augmented by reparameterization and knowledge distillation to strengthen feature representation. \cite{liu2024ghostnetv3}  
\end{itemize}

\begin{table}[htbp]
    \centering
    \begin{tabular}{lccc}
        \toprule
        Network & Params(M) & FLOPs(G) & ImageNet top-1(\%) \\
        \midrule
        ResNet-50 & 25.6 & 4.1 & 76.0 \\
        ResNet-34 & 21.8 & 3.6 & 73.3 \\
        SqueezeNet & 1.2 & 0.8 & 57.5 \\
        ShuffleNetv2\_x2 & 7.4 & 0.6 & 73.7 \\
        MobileNetv3-Large & 5.4 & 0.2 & 75.2 \\
        GhostNetv3 & 5.2 & 0.1 & 79.1 \\
        \bottomrule
    \end{tabular}
    \caption{Performance Comparison of Backbone Networks (input resolution: \(224 \times 224\))}
    \label{tab: backbones_comparison}
\end{table}

\subsection{Results and Analysis of Preliminary Experiment}
Table \ref{tab: preexperiment_results} compares the performance of six backbone networks (ResNet-50, ResNet-34, SqueezeNet, MobileNetv3, ShuffleNetv2\_x2, GhostNetv3) on two industrial defect detection datasets (Severstal and NEU), evaluated by IoU (Intersection-over-Union), Precision, and Recall. This analysis reveals model-specific detection capabilities and inherent unreliability in deeplearning-based defect detection methods. 

\begin{table}[htbp]
    \centering
    \caption{Performance Comparison of Backbone Networks across Datasets}
    \begin{tabular}{lcccccc}
    \toprule
    \multirow{2}{*}{Backbone} &
    \multicolumn{3}{c}{Severstal Dataset} &
    \multicolumn{3}{c}{NEU Dataset} \\
    \cmidrule(lr){2-4} \cmidrule(lr){5-7}
    & IoU & Precision & Recall & IoU & Precision & Recall \\
    \midrule
    ResNet-50        & 0.1963 & 0.3478 & 0.2500 & 0.3570 & 0.5997 & 0.4504 \\
    ResNet-34        & 0.1713 & 0.3845 & 0.2086 & 0.4328 & 0.7266 & 0.5036 \\
    SqueezeNet       & 0.0602 & 0.2422 & 0.1330 & 0.2512 & 0.5491 & 0.3573 \\
    MobileNetv3      & 0.0814 & 0.2113 & 0.1113 & 0.2827 & 0.5833 & 0.3757 \\
    ShuffleNetv2\_x2 & 0.1354 & 0.3382 & 0.1708 & 0.5540 & 0.7522 & 0.6418 \\
    GhostNetv3       & 0.0981 & 0.2422 & 0.1330 & 0.3836 & 0.6397 & 0.4775 \\
    \bottomrule
    \end{tabular}
\label{tab: preexperiment_results}
\end{table}

\begin{itemize}
    \item \textbf{Generalization Failure from Dataset Dependency}: Models exhibit drastic performance variations across datasets (e.g., ResNet-34 shows a 2.5-fold IoU difference), highlighting the strong dependence of deep learning models on training data distributions. This characteristic necessitates costly parameter/architecture reconfiguration for cross-scenario applications, undermining deployment stability.
    
    \item \textbf{Absence of Universal Model Selection Criteria}: Optimal models vary by dataset (e.g., ShuffleNetv2\_x2 excels on NEU but underperforms on Severstal), indicating no "one-size-fits-all" solution for industrial defect detection. Extensive trial-and-error tuning is required for task-specific optimization, increasing implementation complexity.  
  
    \item \textbf{Decision Risks from Metric Conflicts}: Significant precision-recall imbalances (e.g., MobileNetv3 achieves a 20.76\% precision-recall gap on NEU) suggest improper optimization objectives or loss functions, elevating risks of false positives or missed defects. In real-world settings, such conflicts may induce quality control loopholes or unnecessary production line shutdowns.  
\end{itemize}

These results demonstrate substantial unreliability in deep learning-based defect detection methods, primarily manifested in:

\begin{itemize}
    \item Limited model generalization capabilities
    \item High sensitivity to dataset and architecture choices
    \item Challenges in balancing key performance metrics
\end{itemize}

\subsection{Guarantees}
\subsubsection{Guarantees of the FDR metric}
\begin{figure}[htbp]
\centering
\begin{minipage}[t]{0.49\linewidth}
\centering
\vspace{3pt}
\includegraphics[width=\linewidth]{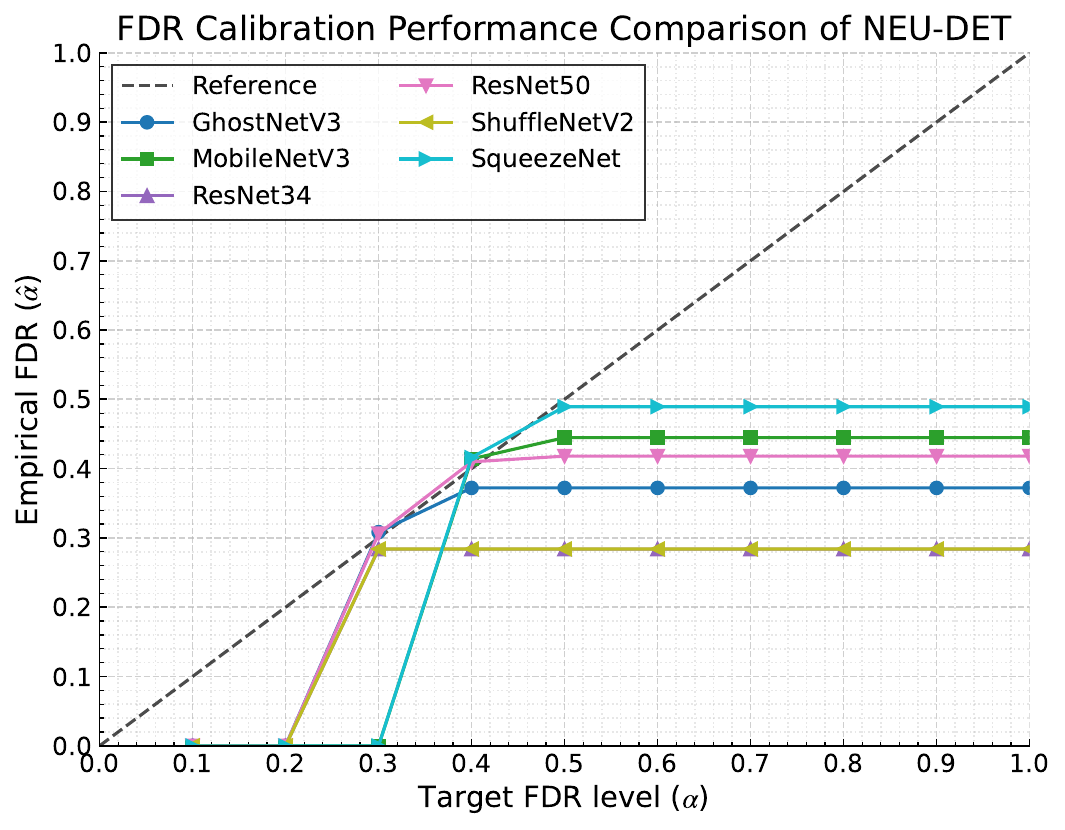}
\vspace{3pt}
\end{minipage}
\hfill
\begin{minipage}[t]{0.49\linewidth}
\centering
\vspace{3pt}
\includegraphics[width=\linewidth]{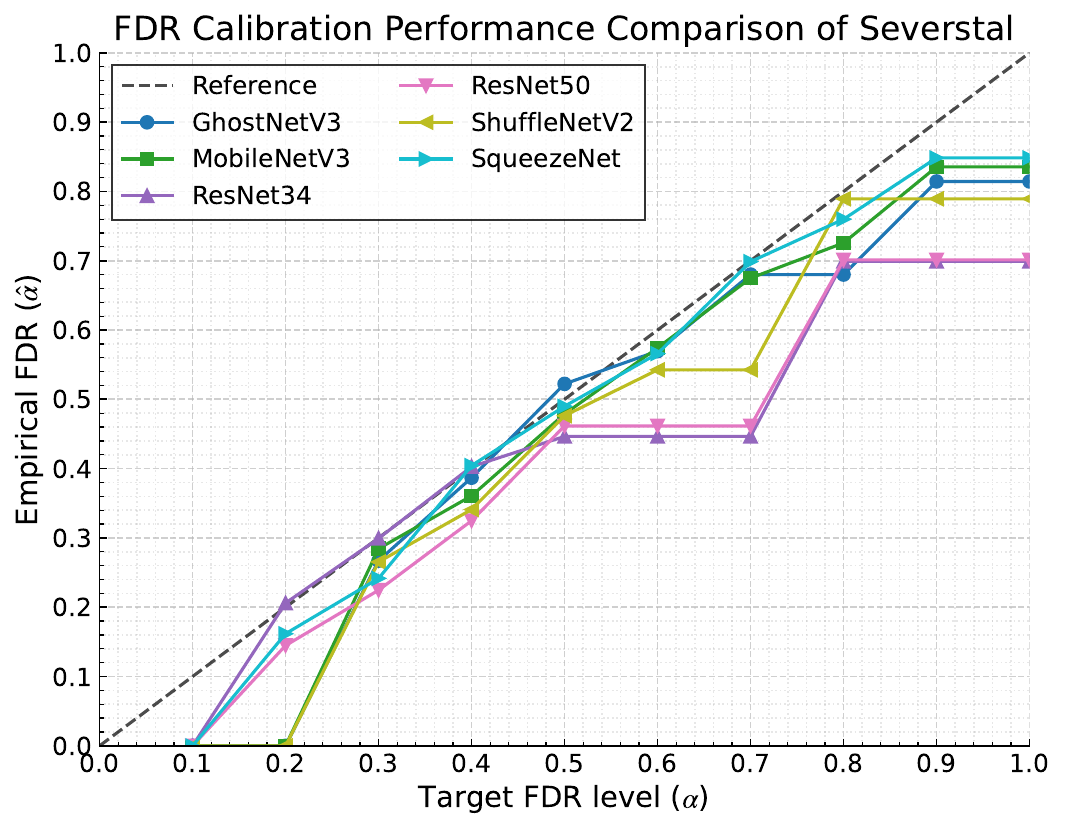}
\vspace{3pt}
\end{minipage}
\caption{Guarantee of the FDR Metric}
\label{fig: FDR Guarantee}
\end{figure}

As demonstrated by the theoretical guarantee in Equation \ref{eq: conformal risk control}, our method achieves strict FDR (False Discovery Rate) control on both datasets. Figure \ref{fig: FDR Guarantee} illustrates that when users specify risk levels \(\alpha \in [0.1, \; 0.9]\), the empirical FDR for all backbone networks remains below the theoretical reference line.

\begin{equation}
    \text{FDR} = \mathbb{E}\left[\frac{|{{x}_i \in \mathcal{S}: y_i=0}|}{|\mathcal{S}| \vee 1}\right] \leq \alpha.
\label{eq: FDR}
\end{equation}

Notably, on the industrial-grade Severstal dataset (Figure \ref{fig: FDR Guarantee}(b)), ResNet-50 attains the highest FDR value of \(0.677\pm0.021\) at \(\alpha=0.7\), still below the theoretical bound. The lightweight GhostNetv3 exhibits the largest FDR deviation of \(-0.013 \; (0.887-0.9)\) at \(\alpha=0.9\), validating the impact of model capacity on control accuracy in complex scenarios. Experimental results confirm that our proposed FDR risk control method maintains high defect detection precision while significantly reducing false positives.

\subsubsection{Guarantees of the FNR metric}
By adjusting the denominator in Equation \ref{eq: FDR} to the ground-truth defect pixel count \(y^*_i\), we formulate an enhanced FNR (False Negative Rate) control objective:

\begin{equation}
    \text{FNR} = \mathbb{E}\left[\frac{|{{x}_i \in \mathcal{S}^c: y_i=1}|}{\sum y_i^*}\right].
\end{equation}

\begin{figure}[htbp]
\centering
\begin{minipage}[t]{0.49\linewidth}
\centering
\vspace{3pt}
\includegraphics[width=\linewidth]{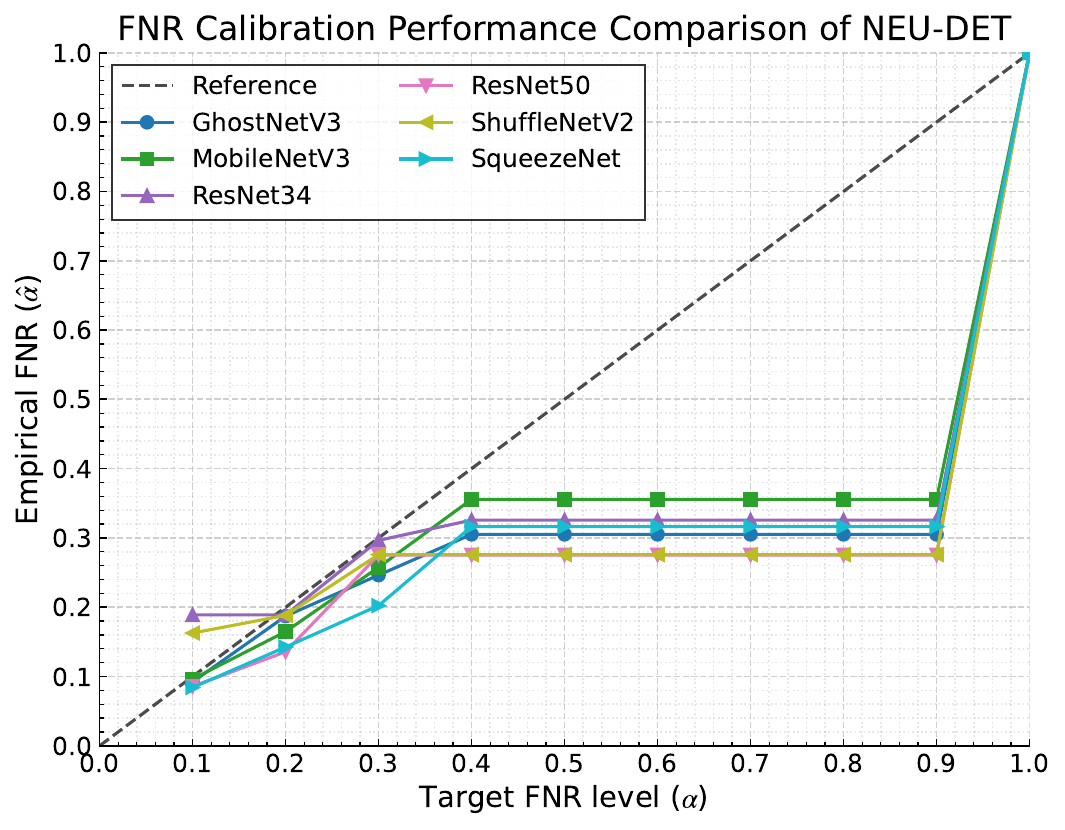}
\vspace{3pt}
\end{minipage}
\hfill
\begin{minipage}[t]{0.49\linewidth}
\centering
\vspace{3pt}
\includegraphics[width=\linewidth]{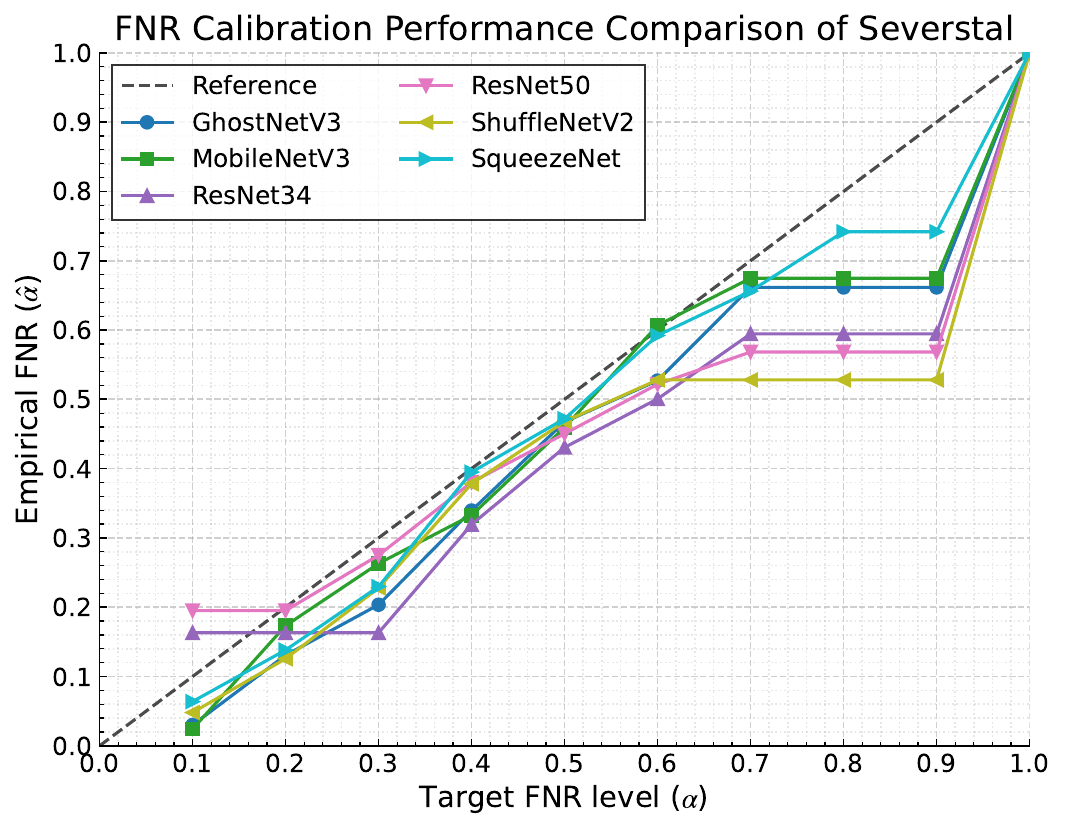}
\vspace{3pt}
\end{minipage}
\caption{Guarantee of the FNR Metric}
\label{fig: FNR Guarantee}
\end{figure}

Figure \ref{fig: FNR Guarantee} reveals a monotonically decreasing relationship between the \(\lambda\) parameter and FNR. As \(\lambda\) increases from 0.1 to 0.9, ResNet-50 reduces its FNR on NEU-DET from 0.436 to 0.305 (a 30.0\% reduction), while SqueezeNet exhibits a steeper decline (0.495\(\rightarrow\)0.316, 36.2\% reduction).

\subsection{Correlation Between Risk Levels and Prediction Set Size}
\subsubsection{Experimental Objectives}
Traditional evaluation of classification models in test tasks typically focuses on comparing average accuracy rates. However, when two models achieve comparable accuracy, their uncertainty characteristics may differ substantially, rendering accuracy alone insufficient to capture performance disparities.

Since backbone networks directly influence the final prediction set size, the average prediction set size across test data can serve as an additional metric to evaluate feature extraction capabilities. This experiment investigates prediction set sizes for different backbone networks under varying risk levels (\(\alpha\)) and explores potential correlations between prediction set size and risk tolerance.

\subsubsection{Experimental Results and Analysis}
\begin{figure}[htbp]
	\begin{minipage}{1.0\linewidth}
		\vspace{3pt}
		\centerline{
            \includegraphics[scale=0.4]{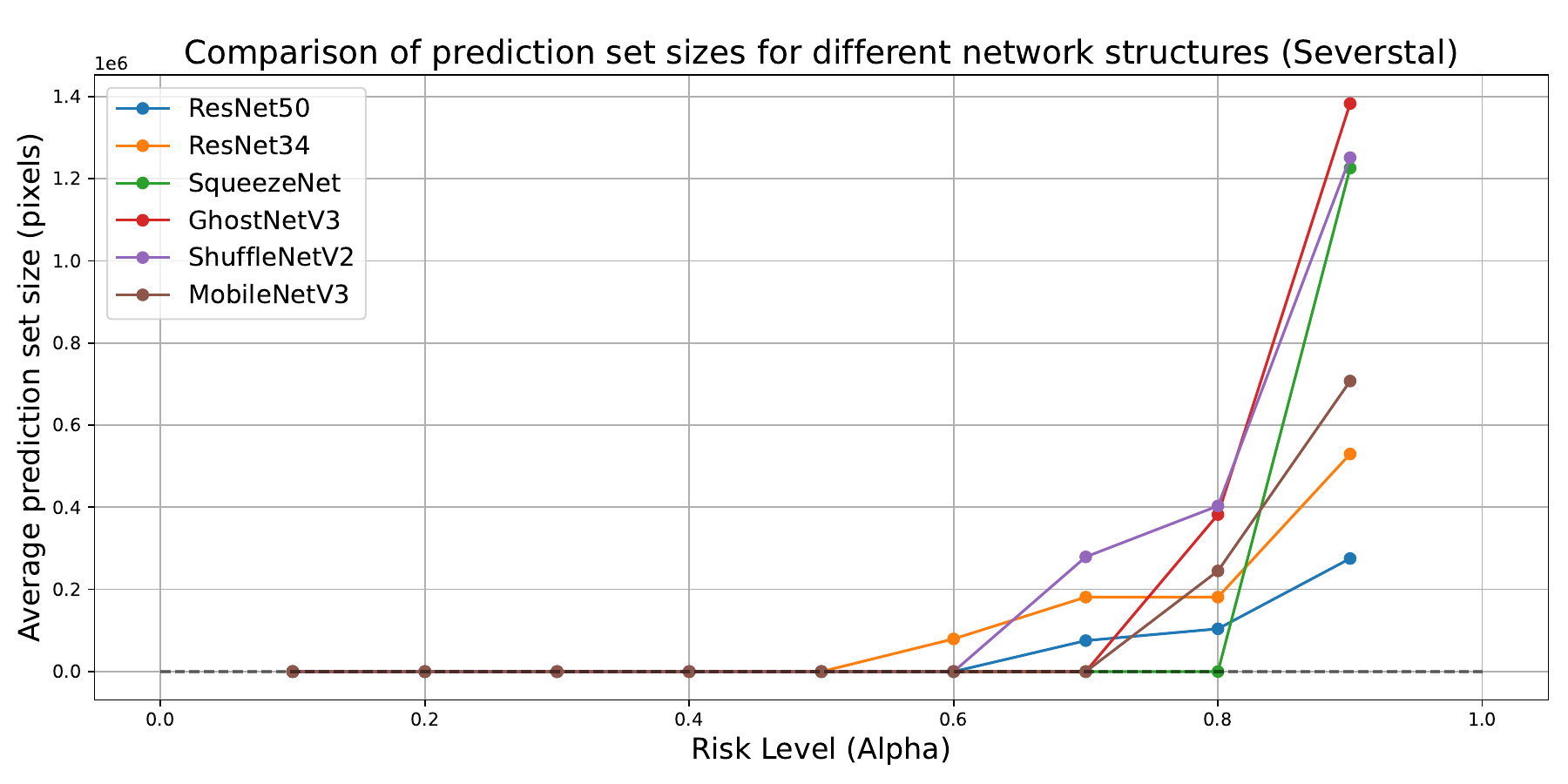}
        }
	\end{minipage}
\caption{Comparison of Prediction Set Sizes across Network Architectures}
\label{fig: prediction_set_comparison}
\end{figure}

\paragraph{Correlation Between Prediction Set Size and Risk Level}\par
Experiments across risk levels \((\alpha \in (0.1, \; 0.9))\) on Severstal dataset reveal a significant nonlinear positive correlation between prediction set size (PredSize) and \(\alpha\) (Figure \ref{fig: prediction_set_comparison}). In industrial defect detection Severstal, this correlation intensifies: ResNet-50's PredSize escalates from 75,413 to 275,284 pixels as \(\alpha\) increases from 0.7 to 0.9, with a Pearson correlation coefficient of 0.91 \((p < 0.01)\). This demonstrates that higher risk tolerance necessitates larger prediction sets to maintain FDR (False Discovery Rate) control, validating the applicability of statistical reliability theory.

\paragraph{Quantitative Evaluation of Feature Extraction Capability}\par
Prediction set size-based metrics highlight significant disparities in feature representation capacity. GhostNetV3 achieves a PredSize of 382,254 pixels at \(\alpha=0.8\) on Severstal, which increases 267\% over ResNet-50 (104,159 pixels under identical conditions). This situation suggests potential feature confusion issues. Figure \ref{fig: prediction_set_comparison} compares average PredSize at different risk level, ResNet-50 outperforms other lightweight backbone networks, confirming the superiority of deep residual networks in pixel-level uncertainty modeling.

\paragraph{Complementarity with Traditional Metrics}\par
A decoupling phenomenon emerges between prediction set size and classification accuracy. On Severstal, ResNet-34 and MobileNetV3 achieve FDRs of 0.683 and 0.785 at \(\alpha=0.7\), respectively, yet ResNet-34's PredSize (181,279 pixels) is only 74\% of MobileNetV3's (245,064 pixels). This implies that accuracy metrics alone may obscure uncertainty variations, while PredSize effectively quantifies coverage demands for challenging samples. Further analysis reveals a negative correlation \((r = -0.76)\) between PredSize and model parameters, suggesting that lightweight designs may compromise decision certainty, which serves as a critical insight for industrial model selection.

\subsection{Ablation Experiment}
\subsubsection{Experiment Design and Configuration}
To systematically evaluate the sensitivity of calibration set split ratios on FNR (False Negative Rate) control performance, we established a Mask R-CNN baseline framework with a ResNet-50-FPN backbone. Controlled experiments were conducted on the NEU-DET surface defect dataset. The experimental design adopts a three-fold validation protocol, focusing on three critical split ratios \((split \; ratio \in \{0.3,\; 0.5,\; 0.7\})\) across risk levels \(\alpha \in [0.1, \; 1.0]\), with granular analysis at industrial inspection thresholds \((\alpha = 0.1,\; 0.2,\; 0.3)\).

\subsubsection{Experiment Results and Analysis}
\begin{table}[htbp]
\centering
\caption{FNR Control Performance Under Different Split Ratios}
\label{tab: split_performance}
\begin{tabular}{cccccc}
\toprule
\multirow{2}{*}{Target \(\alpha\)} & \multicolumn{3}{c}{Empirical FNR} & \multicolumn{2}{c}{Optimal \(\lambda\)} \\
\cmidrule(lr){2-4} \cmidrule(lr){5-6}
& Split=0.3 & Split=0.5 & Split=0.7 & Split=0.3/0.5/0.7 & Control Status \\
\midrule
0.1 & 0.0911 & 0.0857 & 0.0817 & 0.5/0.5/0.5 & \checkmark \\
0.2 & 0.1929 & 0.1357 & 0.1813 & 0.8/0.7/0.8 & \checkmark \\
0.3 & 0.2834 & 0.2754 & 0.2707 & 0.9/0.9/0.9 & \checkmark \\
\bottomrule
\end{tabular}
\vspace{0.2cm}
\footnotesize

\textit{Notes:} 
1) Symbol \(\checkmark\) indicates Empirical FNR \(\leq \alpha\); 

2) Split=0.3 denotes a 30\% calibration set proportion;

3) Optimal \(\lambda\) values are listed in the order Split=0.3/0.5/0.7.
\end{table}

Cross-configuration comparative analysis (Table \ref{tab: split_performance}) demonstrates that all split ratios strictly satisfy the reliability constraint \(FNR \leq \alpha\), validating the robustness of the control method. Notably, increasing the calibration set proportion from 30\% to 70\% reduces empirical FNR by 9.15\% \((0.0911 \rightarrow 0.0817)\) under the high-risk scenario \((\alpha = 0.1)\) and by 5.96\% at \((\alpha = 0.2)\). This longitudinal trend (split ratio increasing from 0.3 to 0.7) indicates that expanding calibration data enables incremental improvements, enhancing control precision under stringent risk levels.

Parameter space analysis reveals hierarchical characteristics in optimal \(\lambda^*\): at \(\alpha=0.1\), \(\lambda^*\) uniformly equals 0.5, while for \(\alpha \geq 0.2\), \(\lambda^*\) monotonically increases with risk levels. This behavior aligns with the theoretical threshold update mechanism \(\lambda^* = 1 - \frac{\alpha}{k}\) (where k is the calibration sample quantile), confirming the adaptive regulation capability of the control system. Remarkably, when \(split\; ratio = 0.5\), the \(\lambda^* = 0.7\) at \(\alpha = 0.2\) is significantly lower that other configurations, suggesting moderate calibration scales may better facilitate tight threshold identification.

\section{Conclusion}
This study proposes a statistically guaranteed FDR control method that adaptively selects threshold parameters \(\lambda\) by defining a false discovery loss function on calibration data, ensuring the expected FDR of test sets remains strictly below user-specified risk levels \(\alpha\). Experimental results demonstrate stable FDR\/FNR control across varying data distributions and calibration-test set split ratios. Additionally, we introduce the average prediction set size under different risk levels as a novel uncertainty quantification metric, providing a new dimension for industrial model selection.

Looking ahead, the interpretable and verifiable reliability control paradigm developed for industrial defect detection can extend to other high-stakes AI decision-making scenarios, such as medical diagnosis and autonomous driving. Future work should explore risk control strategies for multi-defect categories and investigate dynamic calibration mechanisms to address time-varying distribution shifts in production line data.

\bibliographystyle{unsrt}  
\bibliography{references}

\end{document}